\documentclass[letterpaper]{article} % DO NOT CHANGE THIS
\usepackage{aaai25}  % DO NOT CHANGE THIS
\usepackage{times}  % DO NOT CHANGE THIS
\usepackage{helvet}  % DO NOT CHANGE THIS
\usepackage{courier}  % DO NOT CHANGE THIS
\usepackage[hyphens]{url}  % DO NOT CHANGE THIS
\usepackage{graphicx} % DO NOT CHANGE THIS
\usepackage{natbib}  % DO NOT CHANGE THIS AND DO NOT ADD ANY OPTIONS TO IT
\usepackage{caption} % DO NOT CHANGE THIS AND DO NOT ADD ANY OPTIONS TO IT
\usepackage{algorithm}
\usepackage{algorithmic}

\usepackage{subfigure}
\usepackage{amsmath}
\usepackage{amsthm}
\usepackage{arydshln}
\usepackage{booktabs}
\usepackage{multirow}
\usepackage{color}

\usepackage{url}

\usepackage{newfloat}
\usepackage{listings}
\DeclareCaptionStyle{ruled}{labelfont=normalfont,labelsep=colon,strut=off} % DO NOT CHANGE THIS
\floatstyle{ruled}
\newfloat{listing}{tb}{lst}{}
\floatname{listing}{Listing}
\title{Lightweight Contrastive Distilled Hashing for Online Cross-modal Retrieval}
\author{
    Jiaxing Li\textsuperscript{\rm 1},
    Lin Jiang\textsuperscript{\rm 2},
    Zeqi Ma\textsuperscript{\rm 3},
    Kaihang Jiang\textsuperscript{\rm 3},
    Xiaozhao Fang\textsuperscript{\rm 4,\thanks{Corresponding Author: Xiaozhao Fang.}},
    Jie Wen\textsuperscript{\rm 5}
}
\affiliations{
    \textsuperscript{\rm 1}School of Artificial Intelligence, Guangzhou University, Guangzhou 510006, China\\
    \textsuperscript{\rm 2}School of Computer Science, Guangdong Polytechnic Normal University, Guangzhou 510665, China\\
    \textsuperscript{\rm 3}School of Fashion and Textiles, The Hong Kong Polytechic University, Hong Kong SAR, China\\
    \textsuperscript{\rm 4}School of Automation, Guangdong University of Technology, Guangzhou 510006, China\\
    \textsuperscript{\rm 5}Bio-Computing Research Center, Harbin Institute of Technology, Shenzhen 518067, China\\
    jiaxing.li.cs@gmail.com, linn\_jiang@163.com, 22123638r@connect.polyu.hk, jkh1650290810@163.com, xzhfang168@126.com, jiewen\_pr@126.com
    
}

\usepackage{bibentry}
\begin{document}

\maketitle

\begin{abstract}
Deep online cross-modal hashing has gained much attention from researchers recently, as its promising applications with low storage requirement, fast retrieval efficiency and cross modality adaptive, etc.
However, there still exists some technical hurdles that hinder its applications, e.g., 1) how to extract the coexistent semantic relevance of cross-modal data, 2) how to achieve competitive performance when handling the real time data streams, 3) how to transfer the knowledge learned from offline to online training in a lightweight manner.
To address these problems, this paper proposes a lightweight contrastive distilled hashing (LCDH) for cross-modal retrieval, by innovatively bridging the offline and online cross-modal hashing by similarity matrix approximation in a knowledge distillation framework. 
Specifically, in the teacher network, LCDH first extracts the cross-modal features by the contrastive language-image pre-training (CLIP), which are further fed into an attention module for representation enhancement after feature fusion.
Then, the output of the attention module is fed into a FC layer to obtain hash codes for aligning the sizes of similarity matrices for online and offline training.
In the student network, LCDH extracts the visual and textual features by lightweight models, and then the features are fed into a FC layer to generate binary codes.
Finally, by approximating the similarity matrices, the performance of online hashing in the lightweight student network can be enhanced by the supervision of coexistent semantic relevance that is distilled from the teacher network.
Experimental results on three widely used datasets demonstrate that LCDH outperforms some state-of-the-art methods.
\end{abstract}


\section{Introduction}

In the past decades, the volume of multimedia data has explosively grown in an exponential way \cite{CCDR,ijcai2023p398,DRLPP,xu2023untie,ling2023dual,NEURIPS2023_8c64bc3f,chen2024denoising}, especially in some artificial intelligence (AI)-based applications.
Under this circumstance, it is much more difficult to provide the real time retrieval services for the cross-modal data with such huge volume. 
As a powerful model in AI-based applications, large language model (LLM) is benefited from the increasing volume of multimedia data, and has shown its great potentials in representation learning, neural language processing and generation, etc.
Training of LLMs involves a large number of parameters, and thus they can be applied to many complex scenarios for real time prediction or processing \cite{zhu2024large}.
However, the devices that handling the cross-modal retrieval tasks in AI applications may not be powerful in performance, especially for the mobile devices.
Thus, a lightweight retrieval framework supervised by the large training model for multimedia data is with urgent need.

However, the application of lightweight online cross-modal hashing that is supervised by large-scale model training in the knowledge distillation framework, is hindered by some technical hurdles.
Firstly, it is difficult in cross-modal retrieval to extract coexistent semantic relevance of data from different modalities, as there exists huge semantically heterogeneous gaps between different modalities.
Secondly, the feature representation of coexistent semantic relevance on similarity is essential to effectively achieve better performances in hashing-based cross-modal retrieval, but most of the existing methods may fail to adequately obtain the effective representation. 
Finally, bridging the connection between the offline hashing in teacher network and online hashing in the lightweight student network under the knowledge distillation framework is also challenging.

To overcome the above-mentioned technical hurdles, this paper proposes a lightweight contrastive distilled hashing (LCDH) for online cross-modal retrieval.
LCDH follows the research trend that combining traditional methods and modern neural networks to provide a lightweight retrieval framework for AI applications \cite{zhu2024large}.
Specifically, LCDH first fuses the cross-modal features extracted by CLIP, and the fused feature is further fed into an attention module to enhance feature representation in the teacher network.
Then, for aligning the sizes of similarity matrices in offline and online hashing, the output of the attention module is fed into a FC layer.
In the student network, LCDH extracts image and text features by lightweight models, and then the features are fed into a FC layer to obtain binary codes, respectively.
Finally, the performance of online cross-modal hashing in the lightweight student network can be boosted under the supervision of the coexistent semantic relevance which is learned from the teacher network, after similarity approximation by bridging the online and offline similarity matrices.
Main contributions are summarized as follows.

\begin{itemize}
	\item Different from the existing cross-modal hashing methods, the proposed LCDH can effectively extract the coexistent semantic relevance of cross-modal data for the offline cross-modal hashing in teacher network, and then the coexistent semantic relevance is distilled and transferred for guiding the online cross-modal retrieval in the lightweight student network.
	\item To make the lightweight applications of LLMs can be efficiently deployed into the client devices, LCDH first adopts CLIP and attention modules to effectively preserve the coexistent semantic relevance on modalities.
	After the feature extraction and enhancement in the teacher network, the features are fused to obtain discriminative hash codes that can further supervise the online training in lightweight student network by bridging with the offline and online similarity matrices.
	\item Extensive experimental results on three widely used datasets demonstrate that LCDH can deliver the best and stable performance consistently, compared with other state-of-the-art baseline methods.
\end{itemize}

\begin{figure*}[!t]
	\centering
	\includegraphics[width=6.6in]{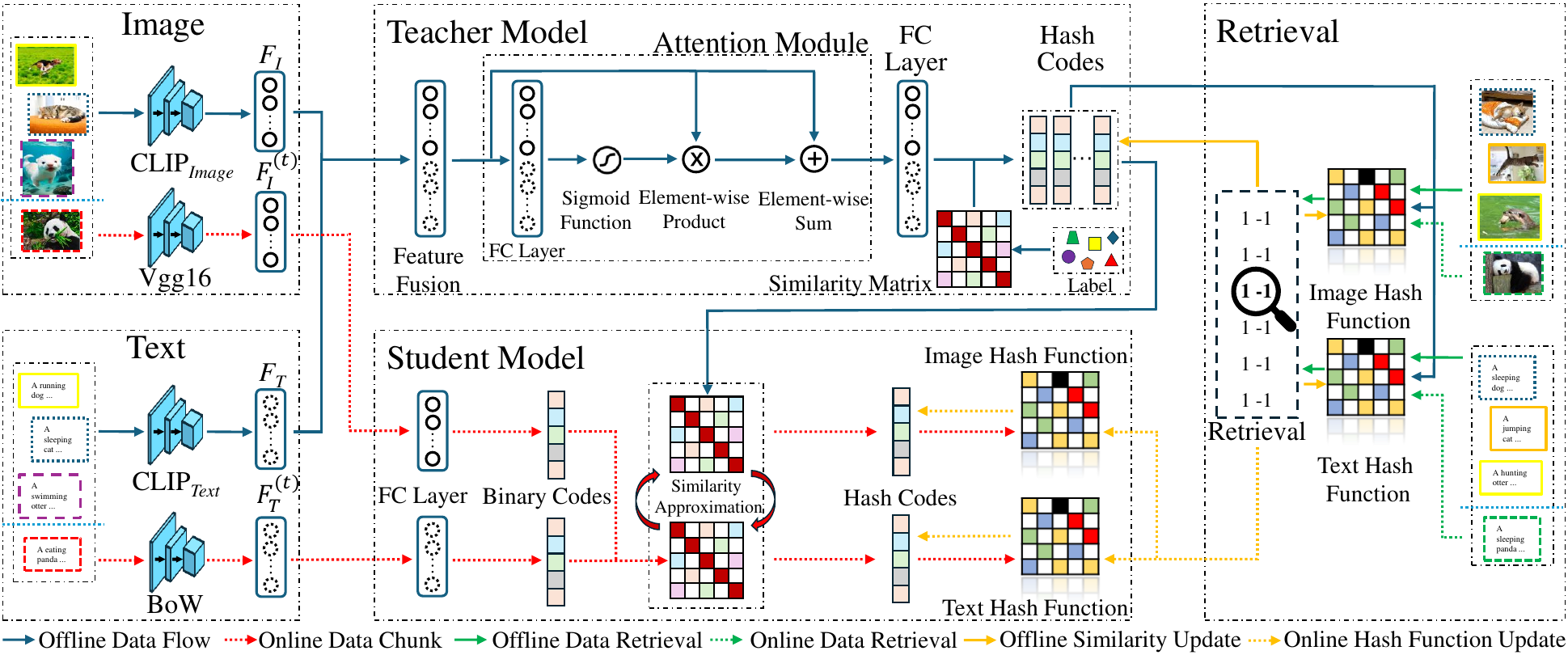}\\
	\centering
	\caption{The framework of the lightweight contrastive distilled hashing (LCDH) for online cross-modal retrieval. 
	}
	\label{framework}
\end{figure*}

\section{Related Works}\label{relatedworks}

\subsection{Offline Cross-modal Hashing}
Offline cross-modal hashing methods attempt to train hash codes and functions with full database in a batch-based manner, which suffers from the limitation that cannot handle the real time data chunks \cite{huang2023two}.
For example, deep cross-modal hashing (DCMH) adopts the end-to-end strategy to jointly learn hash codes and functions for coss-modal data \cite{DCMH}.
Self-supervised adversarial hashing (SSAH) leverages two adversarial-based models to maximize correlations on semantic between modalities \cite{SSAH}.
Unsupervised knowledge distillation (UKD) hashing constructs similarity matrix in the unsupervised teacher network for guiding the hash codes and functions learning in the supervised student network \cite{UKD}.
Deep cross-modal hashing with hash functions and unified hash code jointly learning (DCUCH) generates hash codes and learns hash functions asymmetrically \cite{DCHUC}.
Deep adaptive enhanced hashing (DAEH) uses adaptive optimization method and discriminative similarity measurement to learn hash functions \cite{DAEH}.
Semi supervised knowledge distillation for cross-modal hashing (SKDCH) learns similarity as knowledge in a semi-supervised method in teacher network to guide the supervised cross-modal hashing in student network \cite{SKDCH}.
Coding self-representative and label-relaxed hashing (CSLRH) enhance the class-specific feature representations and the semantic discriminability, while reduce the noise interference \cite{jiang2024coding}.
CLIP based knowledge distillation hashing (CKDH) leverages a knowledge distillation framework to transfer the deep semantic similarity learned by CLIP to the lightweight corss-modal hashing in the student network \cite{CKDH}.

\subsection{Online Cross-modal Hashing}
Unlike offline cross-modal hashing methods, online cross-modal hashing methods train hash codes and functions in a real-time updating manner, where hash functions are updated by hash codes trained by the continually coming data chunks.
For example, hadamard code book based online hashing (HCOH) leverages hadamard code book to generate hadamard matrix for online hash codes and functions learning \cite{HCOH}.
Online latent semantic hashing (OLSH) maps discrete labels to a continuous latent semantic space which is bridged with binary codes for learning hash codes \cite{OLSH}.
Online collective matrix factorization hashing (OCMFH) uses the collective matrix factorization method to decompose new data chunk's features and existing features into a common latent space that is generated by the features and hash function simultaneously \cite{OCMFH}.
Label embedding online hashing (LEMON) leverages similarity factorization hashing method for addressing the online hashing retrieval problem, by reconstructing the similarity matrices using label information for newly coming online data chunks \cite{LEMON}.
Supervised discrete online hashing (SDOH) embeds semantic label into the common latent space for parallel calculating the common representation of the newly coming data from multiple modalities, and then generates hash codes by the continuous common latent representation \cite{SDOH}.
Random online hashing (ROH) proposes a linear bridging strategy to simplify the similarity factorization problem into a linear optimization problem, and then proposes a MED embedding methods to learn features and preserve significant semantic information into hash codes \cite{ROH}.

\section{The Proposed Method}

\subsection{Problem Statement}
Assuming that there have $n$ pairs of image-text instances in the training datasets $\mathcal{O}=\{o_{i}\}_{i=1}^{n}$.
Let $v_{j}$ be the $j$-th image instance and $b_{j}$ be the $j$-th text instance, each pair of instances in $\mathcal{O}$ can be represented as $o_{j}=\{v_{j},b_{j}\}$.
The label matrix for $\mathcal{O}$ can be denoted as $L \in  \{0,1\}^{n \times c}$, and $c$ is the number of categories.
The deep semantic features that are extracted from the image feature encoder in the offline manner can be denoted as $F_{I} \in \Re^{n \times d_I}$, while the deep semantic features that are extracted from the text feature encoder in the offline manner can be denoted as $F_{T} \in \Re^{n \times d_T}$.
Similarly, the deep semantic features that are extracted from the image and text feature encoders at round $t$ in the online manner can be denoted as $F_{I}^{(t)} \in \Re^{n \times d_I}$ and $F_{T}^{(t)} \in \Re^{n \times d_T}$, respectively.
In addition, $d_I$ is the dimension of extracted image features, and $d_T$ is the dimension of extracted text features.
$||\cdot||_{F}^{2}$ is the Frobenius norm, in which $||M||_{F}^{2} = Tr(M^{T}M)$ and $Tr(M)$ is the trace of matrix $M$.

Moreover, training set $\mathcal{O}$ is divided into two parts, in which former $n_o$ pairs instances are utilized for offline training in the teacher network and the rest $n_t$ pairs instances are utilized for online training in the student network.
That is, $v_{*}=\{v_{*}^{o}, v_{*}^{t}\}$, $b_{*}=\{b_{*}^{o}, b_{*}^{t}\}$, and $n = n_o + n_t$.
Similarly, the label matrices for instances in offline cross-modal hashing is $L \in \{0,1\}^{n_o \times c}$, and in online cross-modal hashing is $L^{(t)} \in \{0,1\}^{n_t \times c}$.
%The hash codes for image and text instances in offline cross-modal hashing can be represented as $V=\{v_{*,i}^{o}\}_{i=1}^{n_o} \in \{-1,+1\}^{n_o \times r}$ and $B=\{b_{*,i}^{o}\}_{i=1}^{n_o} \in \{-1,+1\}^{n_o \times r}$, where $r$ is the length of hash codes.
After deep feature extraction, the deep semantic features for image and text instances in offline cross-modal hashing are fused and fed into an attention module for further cross-modal semantic preservation.
The output of the attention module is fed into a fully-connected (FC) layer, and then the hash codes $B' \in \{-1,+1\}^{n_t \times r}$ is generated for constructing the similarity matrix.
Notably, $r$ is the length of hash codes.
The hash codes for image and text instances in online cross-modal hashing can be represented as $V^{(t)}=\{v_{*,i}^{t}\}_{i=1}^{n_{t}} \in \{-1,+1\}^{n_{t} \times r}$ and $B^{(t)}=\{b_{*,i}^{t}\}_{i=1}^{n_{t}} \in \{-1,+1\}^{n_{t} \times r}$, where $n_{t}$ is the size of new coming data chunk.

\subsection{Feature Extraction}
The CLIP model takes cross-modal data as input to extract deep semantic features for offline cross-modal hashing in teacher network, while the Vgg16 and BoW models are utilized for extracting deep semantic features for online cross-modal hashing in lightweight student network.
Notably, to make the semantic similarity preserved in the teacher network can supervise the online training in student network, the length of output from the FC layer is set to $n_t$.
As such, the size of similarity matrix $S$ generated by hash codes learned in the teacher network and similarity matrix $S^{(t)}$ updated in the student network is same (i.e., $n_{t} \times n_{t}$).

\textbf{Offline feature extraction:} 
In the teacher network, the CLIP model takes $\{o_{i}\}_{i=1}^{n_o}$ as input, and outputs $F_{I}=\{f_{i}^{1},f_{i}^{2},\dots,f_{i}^{n_o}\} \in \Re^{n_o \times d_{I}}$ and $F_{T}=\{f_{t}^{1},f_{t}^{2},\dots,f_{t}^{n_o}\} \in \Re^{n_o \times d_{T}}$, where $d_{I}$ and $d_{T}$ are the dimensions of $F_{I}$ and $F_{T}$, respectively.
Specifically, $f_{i}^{j} = CLIP_{Image}(v_{j})$ and $f_{t}^{j} = CLIP_{Text}(b_{j})$, where $j \in [1,n_o]$.
Then, the fusion of features $F_{I}$ and $F_{T}$ is input into the attention module as $F_{f}$.
That is, $F_{f} = \{f_{i}^{1}f_{t}^{1}, f_{i}^{2}f_{t}^{2}, \dots, f_{i}^{n_o}f_{t}^{n_o}\}$.
%\textbf{Attentive Feature Extraction:} 
The deep features $F_f$ is then fed into a FC layer to generate binary code $B \in \{-1,+1\}^{n_o \times r}$.
The filter process in attention module can be defined as follows:
\begin{equation}
	M_{f}=Sigmoid(WF_{f}+D),
\end{equation}
where $M_{f} \in \Re^{n_o \times d}$ is the mask weight matrix and $d$ is the dimension of the fused feature.
Moreover, $D$ and $W$ are the parameters for the FC layer, and $Sigmoid$ is the activation function.
The output of attention module is
\begin{equation}
	A_{f} = F_{f} + M_{f} \odot F_{f},
\end{equation}
where $\odot$ indicates element wise product and $A_{f} \in \Re^{n_o \times d}$.
Finally, $A_{f}$ is fed into the FC layer for generating hash codes (i.e., $B' \in \{-1,+1\}^{n_t \times r}$) which are well preserved the modality-specific similarity.
As such, $B'$ can be further utilized for constructing the similarity matrix $S$ to supervise the training in online cross-modal hashing of student network.

\textbf{Online Feature Extraction:} In the student network, we adopt lightweight feature extraction models for online training.
Specifically, we utilize Vgg16 model to extract deep semantic feature $F_{I}^{(t)}$ for image instances, and we utilize BoW model to extract deep semantic feature $F_{T}^{(t)}$ for text instances.
That is, $F_{I}^{(t)} = Enc_{Vgg16}(\{v_{j}\}_{j=1}^{n_t})$ and $F_{T}^{(t)} = Enc_{BoW}(\{b_{j}\}_{j=1}^{n_t})$, where $Enc_{*}(\cdot)$ is the encoder.
After feature extracting, $F_{I}^{(t)}$ and $F_{T}^{(t)}$ are fed into the FC layer for further generating binary codes.

\subsection{Similarity Preserving}

\textbf{Similarity preservation in teacher network:} 
In the teacher network, the output of FC layer is further supervised by the label information in similarity matrix.
Following \cite{DBLP:conf/aaai/ZhangL14}, we randomly select $n_t$ instances from offline training set $\{v_{*}^{o}, b_{*}^{o}\}$ to initialize the similarity matrix as follows:
\begin{equation}\label{S}
	\begin{split}
		S = 2GG^{T} - \mathbf{1}\mathbf{1}^{T},
	\end{split}
\end{equation}
where $L'$ is the label matrix of the selected $n_t$ instances, $G \in \{0,1\}^{n_t \times c}$ is generated by label matrix $L' \in \{0,1\}^{n_t \times c}$ and $G=L'/||L'||_{F}^{2}$.
Then, the $S$ in the offline cross-modal hashing of teacher network can be constructed as follows.
\begin{equation}\label{LT}
	\begin{split}
		& \mathcal{L}_{T} = \mathop{\min}_{B'} ||rS - B'^{T}B'||_{F}^{2},\\
	\end{split}
\end{equation}

\textbf{Similarity preservation in student network:} 
Normally, in the online cross-modal hashing of student network, the semantic similarity between cross-modal data can be embedded into binary codes as follows \cite{6247912}:
\begin{equation}
	\begin{split}
		& \quad\quad\quad \min_{V^{(t)},B^{(t)}} ||rS^{(t)} - V^{(t)}B^{(t)^{T}}||_{F}^{2},\\
	\end{split}
\end{equation}
where $S^{(t)} \in \{-1,+1\}^{n_t \times n_t}$ is the semantic similarity matrix. 
For the $i$-th and $j$-th online instances, if they share at least 1 common label $S_{i,j}^{(t)}=1$, otherwise $S_{i,j}^{(t)}=0$.
However, this method may fail to preserve the fine-grained semantic similarity for multiple label data.
At the same time, this method may also lead to large overheads on storage and computation during training, which goes against the original intention of hashing-based cross-modal retrieval.
To overcome the above issues and make $S^{(t)}$ available for online scenarios, this paper adopts logistic regression to preserve the similarity for cross-modal data.
Let $g(\Omega_{i,j}^{(t)}) = G_{i,j}^{(t)} = \frac{1}{1+e^{-\Omega_{i,j}^{(t)}}}$, where $\Omega_{i,j}^{(t)} = \frac{\omega}{r}B_{i,*}^{(t)}V_{j,*}^{(t)}$ and $\omega$ is a scaling factor of tuning.
Then, we can define the likelihood probability function for $S^{(t)}$ as follows:
\begin{equation}
	\begin{split}
		p(S_{i,j}^{(t)}|B^{(t)})=
		\begin{cases}
			G_{i,j}^{(t)}, & \text{if $S_{i,j}^{(t)}=1$},\\
			1 - G_{i,j}^{(t)}, & \text{otherwise.}
		\end{cases}
	\end{split}
\end{equation}
where the similarity matrix $S^{(t)}$ is constructed by label matrix (i.e., $S^{(t)}=L^{(t)}L^{(t)^{T}}$ and $L^{(t)} \in \{0,1\}^{n_{t} \times c}$) and the value of elements in $S^{(t)}$ is binary as follows:
\begin{equation}
	S_{i,j}^{(t)}=
	\begin{cases}
		1, & \text{if the $i$-th and $j$-th samples are similar,}\\
		0, & \text{otherwise.}
	\end{cases}
\end{equation}
To obtain the maximum likelihood estimation, a log-likelihood similarity preserving term is adopted as follows:
\begin{equation}
	\begin{split}
		&\mathop{\log{L}} = \sum_{i=1}^{n_t}\sum_{j=1}^{n_t} \log p(S_{i,j}^{(t)}|B^{(t)},V^{(t)}).\\
	\end{split}
\end{equation}
Finally, the maximum log-likelihood similarity preserving problem can be defined as follows:
\begin{equation}\label{Llog}
	\begin{split}
		& \mathop{\max}_{S^{(t)}} \mathcal{L}_{log} = \sum_{i=1}^{n_t}\sum_{j=1}^{n_t} [S_{i,j}^{(t)}\Omega_{i,j}^{(t)} - \log(1+e^{\Omega_{i,j}^{(t)}})].\\
	\end{split}
\end{equation}

\subsection{Knowledge Distillation}
After offline cross-modal hashing in teacher network, the semantic similarity is first initialized by Eq. (\ref{S}) and then constructed by Eq. (\ref{LT}).
As such, the deep semantic cross-modal similarity extracted by CLIP and attention module in teacher network can be well preserved by Eq. (\ref{Llog}), and further transferred to the online training in lightweight student network by the knowledge distillation framework in LCDH.
In the lightweight student network, the image and text features are extracted by Vgg16 and BoW models from the online training set in a lightweight manner, respectively.
Then, the hash codes and functions are updated by the maximum log-likelihood similarity preservation.
Meanwhile, the similarity matrix $S^{(t)}$ is updated by $S$ as follows:

\begin{equation}\label{LKD}
	\begin{split}
		& \quad\quad\quad \mathop{\min}_{S^{(t)}} \mathcal{L}_{KD} = ||S - S^{(t)}||_{F}^{2}.\\
	\end{split}
\end{equation}
As such, the deep semantic similarity extracted from CLIP and attention module is distilled and transferred to $S^{(t)}$, which guides the online cross-modal hashing in the lightweight student network.

\subsection{Objective Function}
By integrating Eq. (\ref{LT}), (\ref{Llog}) and (\ref{LKD}), the objective function of LCDH can be formulated as follows:

\begin{equation}\label{objectiveFunction}
	\begin{split}
		&\quad\quad \mathop{\min}_{B',V^{(t)},B^{(t)},S^{(t)}} \mathcal{L} = - \mathcal{L}_{log} + \lambda_1 \mathcal{L}_{T} + \lambda_2 \mathcal{L}_{KD}, \\
	\end{split}
\end{equation}
where $\lambda_1$ and $\lambda_2$ are the balancing parameters.

\section{Experiments}
\begin{table*}[!htbp]
	\centering
	\setlength{\tabcolsep}{1mm}
	\fontsize{9}{9}\selectfont
	\begin{tabular}{ cccccccccccccc}
		\toprule[1.0pt]
		\multirow{2}{*}{Task}& \multirow{2}{*}{Method} &  \multicolumn{4}{c}{MIRFlickr-25K}   & \multicolumn{4}{c}{IAPR TC-12}    & \multicolumn{4}{c}{NUS-WIDE}
		\\ \cmidrule(r){3-6}  \cmidrule(r){7-10}  \cmidrule(r){11-14}  
		&                         & 16 bits          & 32 bits          & 64 bits          & 128 bits         & 16 bits          & 32 bits          & 64 bits          & 128 bits         & 16 bits          & 32 bits          & 64 bits          & 128 bits         \\ \cmidrule(r){3-6}  \cmidrule(r){7-10}  \cmidrule(r){11-14}  
		\multirow{8}{*}{\begin{tabular}[c]{@{}l@{}}\\\\ \\ \\ I$\rightarrow$T \\ \\  \end{tabular}}  
		%		\specialrule{0em}{2pt}{1pt}
		%		\cdashline{2-14}[3pt/2pt]
		%		\specialrule{0em}{2pt}{1pt}
		& DCMH      & 0.7376   & 0.7466  & 0.7469  & 0.7547  & 0.4526 & 0.4732 & 0.4844 & 0.4983    & 0.5783   & 0.5921  & 0.5984  & 0.6097  \\
		& SSAH      & 0.7654   & 0.7728  & 0.7809  & 0.7892  & 0.5381 & 0.5668 & 0.5862 & 0.5992    & 0.6036   & 0.6187  & 0.6404  & 0.6428  \\
		& UKD     & 0.7138   & 0.7184  & 0.7255  & 0.7201   & 0.4812 & 0.4920 & 0.4941 & 0.5027    & 0.6149   & 0.6378  & 0.6386  & 0.6454 \\	
		& DCHUC   & 0.7576   & 0.7557  & 0.7611  & 0.7495  & 0.5545 & 0.6023 & 0.6257 & 0.6484    & 0.7501   & 0.7759  & 0.7976  & 0.7953 \\
		& DAEH     & 0.7802   & 0.7916  & 0.7984  & 0.8027  &0.5428 &0.5796 &0.6011 &0.6087     & 0.7299   & 0.7534  & 0.7708  & 0.7769 \\
		& SKDCH     & 0.7216   & 0.7335  & 0.7382  & 0.7403  &0.5421 &0.5714 &0.5998 &0.6103    & 0.7314   & 0.7444  & 0.7642  & 0.7701 \\
		\specialrule{0em}{2pt}{1pt}
		\cdashline{2-14}[3pt/2pt]
		\specialrule{0em}{2pt}{1pt}
		& OLSH    & 0.5784   & 0.5918  & 0.5981  & 0.5961  & 0.3465   & 0.3519  & 0.3619  & 0.3665  & 0.5312   & 0.5257  & 0.5295  & 0.5354    \\
		& OCMFH    & 0.5860   & 0.5881  & 0.5894  & 0.5875  & 0.3056   & 0.3024  & 0.2961  & 0.3005  & 0.3761   & 0.3778  & 0.3821  & 0.3866     \\
		& LEMON    & 0.7226   & 0.7408  & 0.7402  & 0.7437  & 0.4879   & 0.5101  & 0.5252  & 0.5318  & 0.6535   & 0.6600  & 0.6594  & 0.6702     \\
		& SDOH    & 0.7101   & 0.7268  & 0.7332  & 0.7330  & 0.4787   & 0.4905  & 0.5020  & 0.5084  & 0.6449   & 0.6582  & 0.6716  & 0.6780     \\
		& ROH    & 0.7346   & 0.7525  & 0.7585  & 0.7626  & 0.4842   & 0.5094  & 0.5347  & 0.5471  & 0.6701   & 0.6854  & 0.6928  & 0.7001     \\
		\cmidrule(r){2-6}  \cmidrule(r){7-10}  \cmidrule(r){11-14} 
		& \textbf{LCDH}      & \textbf{0.7898} & \textbf{0.7992} & \textbf{0.8034} & \textbf{0.8041}  &\textbf{0.5656} &\textbf{0.6093} &\textbf{0.6301} &\textbf{0.6499}  & \textbf{0.7564} & \textbf{0.7803}  & \textbf{0.8007} & \textbf{0.8015}\\
		\midrule[1.0pt]
		\multirow{9}{*}{\begin{tabular}[c]{@{}l@{}}\\ \\ \\ \\ T$\rightarrow$I \\ \\ \end{tabular}}  \\
		& DCMH      & 0.7628   & 0.7733  & 0.7792  & 0.7857  & 0.5185 & 0.5378 & 0.5468 & 0.5596     & 0.6277   & 0.6413  & 0.6491  & 0.6503 \\
		& SSAH      & 0.7768   & 0.7847  & 0.7811  & 0.7823  & 0.5392 & 0.5646 & 0.5873 & 0.5998     & 0.6140   & 0.6285  & 0.6306  & 0.6324 \\
		& UKD     & 0.7156   & 0.7166  & 0.7214  & 0.7190  & 0.4886 & 0.5017 & 0.5145 & 0.5172     & 0.6304   & 0.6562  & 0.6573  & 0.6639 \\	
		& DCHUC     & 0.7680   & 0.7791  & 0.7753  & 0.7681  & 0.5393 & 0.6039 & 0.6431 & 0.6734     & 0.7051   & 0.7243  & 0.7479  & 0.7453 \\
		& DAEH    & 0.7597  & 0.7667  & 0.7741  & 0.7813  &0.5132 &0.5558 &0.5799 &0.5905     & 0.7130   & 0.7353  & 0.7455  & 0.7501 \\
		& SKDCH    & 0.7814   & 0.7985  & 0.8067  & 0.7982  &0.5404 &0.6021 &0.6398 &0.6701    & 0.7228   & 0.7404  & 0.7519  & 0.7583 \\
		\specialrule{0em}{2pt}{1pt}
		\cdashline{2-14}[3pt/2pt]
		\specialrule{0em}{2pt}{1pt}
		& OLSH    & 0.5988   & 0.5996  & 0.6011  & 0.6031  & 0.3521   & 0.3668  & 0.3625  & 0.3697  & 0.5604   & 0.5638  & 0.5880  & 0.5841     \\
		& OCMFH    & 0.5829   & 0.6094  & 0.6038  & 0.6024  & 0.3311   & 0.3173  & 0.3286  & 0.3203  & 0.4092   & 0.4175  & 0.4282  & 0.4330     \\
		& LEMON    & 0.8137   & 0.8270  & 0.8333  & 0.8304  & 0.5956   & 0.6185  & 0.6507  & 0.6687  & 0.7772   & 0.7850  & 0.8053  & 0.8018     \\
		& SDOH    & 0.7605   & 0.7961  & 0.7981  & 0.8032  & 0.5644   & 0.5918  & 0.6223  & 0.6441  & 0.7734   & 0.7859  & 0.7982  & 0.7991     \\
		& ROH    & 0.8194   & 0.8362  & 0.8443  & 0.8481  & 0.5765   & 0.6147  & 0.6535  & 0.6795  & 0.7940   & 0.8092  & 0.8162  & 0.8264     \\
		\cmidrule(r){2-6}  \cmidrule(r){7-10}  \cmidrule(r){11-14} 
		& \textbf{LCDH}    & \textbf{0.8223} & \textbf{0.8480} & \textbf{0.8556} & \textbf{0.8574}  &\textbf{0.5987} &\textbf{0.6229} &\textbf{0.6574} &\textbf{0.6810}  & \textbf{0.8018} & \textbf{0.8116}  & \textbf{0.8268} & \textbf{0.8311}\\
		\bottomrule[1.0pt]
	\end{tabular}
	\caption{ mAP results of LCDH and baseline methods on MIRFlickr-25K, IAPR TC-12 and NUS-WIDE.}
	\label{table_MAP}
\end{table*}

\begin{figure*}[!t]
	\centering
	\includegraphics[width=6.2in]{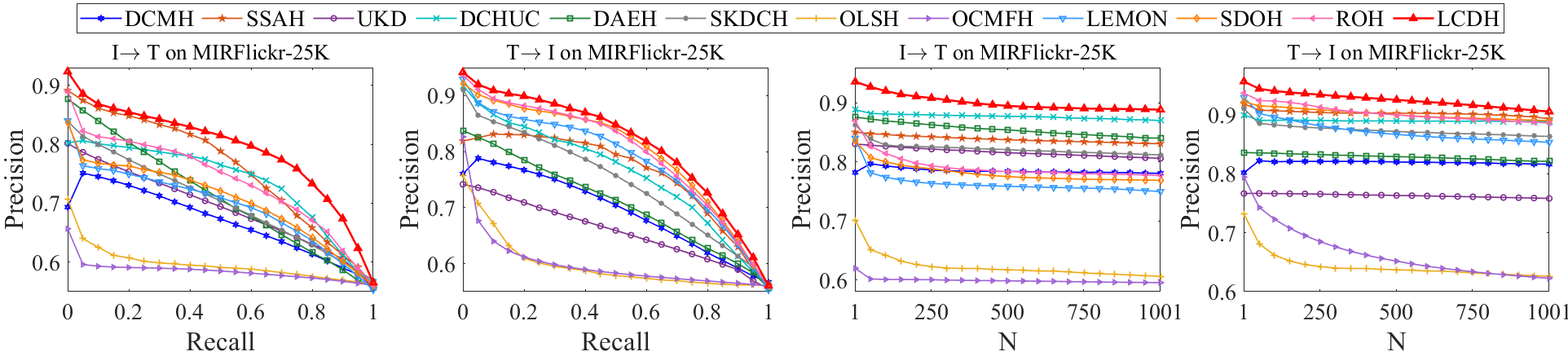}\\
	\centering
	\caption{Precision-recall and Top-N precision curves of LCDH and baselines on MIRFlickr-25K with 128 bits hash codes.}
	\label{mirflickr_PR_TopN}
\end{figure*}

\subsection{Datasets}
\textbf{MIRFlickr-25K} consists of 25000 image-text pairs that annotated with 1 or more of 24 categories from the Flickr website \cite{mirflickr}.
Following \cite{7299011}, in the experiment, we only selected the instances with frequent text tags that appearing more than 20 times.
As such, 16738 pairs of them were remained, in which 2000 pairs of them were selected as the query set and rest of them were selected for the training and retrieval sets.

\textbf{IAPR TC-12} consists of 20000 pairs of image-text instances annotated with 255 categories \cite{iaprtc}.
For each pair of them, the image is represented as a 512-dimensional GIST vector, and the text is represented as a 2912-dimensional BoW vector.
In the experiment, only 2000 pairs were randomly selected as the query set, and the rest were selected for the training and retrieval sets.

\textbf{NUS-WIDE} consists of 269648 pairs of image-text instances annotated with 81 categories \cite{nuswide}.
In the experiment, pairs of instances that annotated with 10 most frequent categories in the original 81 categories were chosen.
Thus, there were 186577 labeled pairs were selected, in which only 2100 pairs were treated as query set and the rest were treated as the training and retrieval sets.

\subsection{Evaluation Metrics}

The performances of two tasks on cross-modal retrieval are evaluated in this section, which are text to image (T$\rightarrow$I) and image to text (I$\rightarrow$T) tasks, respectively.
As a widely accepted metric for the performance evaluation in the cross-modal retrieval, mean average precision (mAP) is adopted as one of the evaluation metrics in this paper.
mAP is actually the mean of average precision, which can verify the performance of the model more comprehensively and accurately \cite{wang2016comprehensive}.
Moreover, the precision recall and top-N precision are also the widely accepted performance evaluation metrics for cross-modal retrieval tasks.

\subsection{Baseline Methods and Implementation Details}
To verify the superior performances on cross-modal retrieval of LCDH, we compared LCDH with 6 offline cross-modal hashing methods (i.e., DCMH$_{2017}$, AGAH$_{2019}$, UKD$_{2020}$, DCHUC$_{2022}$, DAEH$_{2022}$ and SKDCH$_{2023}$, and 5 online cross-modal hashing methods (i.e., OLSH$_{2019}$, OCMFH$_{2020}$, LEMON$_{2020}$, SDOH$_{2022}$, ROH$_{2023}$).
Notably, UKD$_{2020}$ and SKDCH$_{2023}$ are the knowledge distillation-based hashing methods.
All experiments were conducted on a workstation that running Windows 10 professional operation system with AMD Ryzen9 5900X CPU and NVIDIA GeForce RTX3090 GPU.
The versions of CUDA is 11.3, of python is 3.9.8 and of pytorch is 1.10.1.

\begin{figure*}[!t]
	\centering
	\includegraphics[width=6.2in]{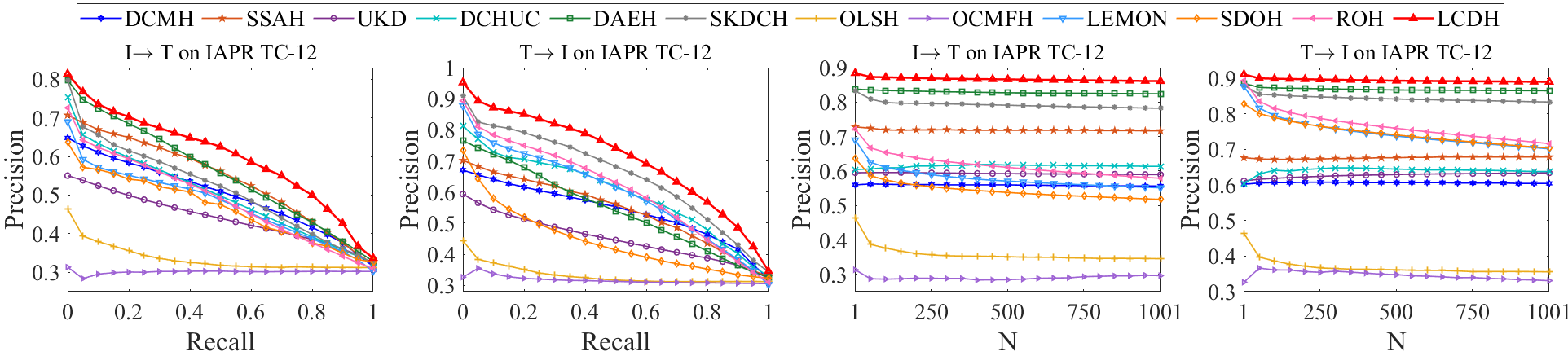}\\
	\centering
	\caption{Precision-recall and Top-N precision curves of LCDH and baselines on IAPR TC-12 with 128 bits hash codes.}
	\label{iaprtc_PR_TopN}
\end{figure*}

\begin{figure*}[!t]
	\centering
	\includegraphics[width=6.2in]{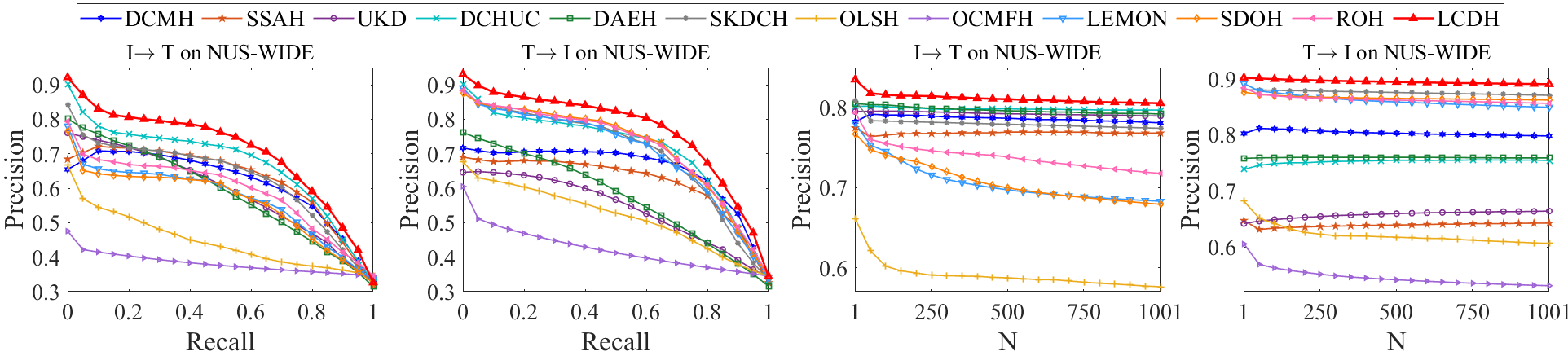}\\
	\centering
	\caption{Precision-recall and Top-N precision curves of LCDH and baselines on NUS-WIDE with 128 bits hash codes.}
	\label{nuswide_PR_TopN}
\end{figure*}

\subsection{Results on Datasets}
As shown in Table \ref{table_MAP}, the mAP results of the proposed method outperform other baseline methods in terms of I$\rightarrow$T and T$\rightarrow$I tasks on MIRFlickr-25K, IAPR TC-12 and NUS-WIDE.
Normally, the offline cross-modal hashing baseline methods will outperform the online ones. 
This is mainly because the offline methods learns hash codes and functions in two steps with full database, while the online methods learns hash codes and updates hash functions with data chunks of partial database.
The results of the proposed LCDH and baseline methods on mAP, precision recall curve and top-N precision curve are as follows.

\subsubsection{Results on mAP}
From Table \ref{table_MAP}, we can have the following observations and analyses of mAP results.

\begin{itemize}
	\item LCDH outperforms knowledge distillation-based UKD and SKDCH methods, verifying that both knowledge distillation framework and features extraction modules are effective and have considerable contributions to the performance of LCDH.
	The main reason may be that the feature extraction by CLIP and feature representation enhanced by the attention module in LCDH are more effective than that in UKD and SKDCH.
	\item All methods deliver outstanding results on mAP for MIRFlickr-25K and NUS-WIDE, but ordinary one for IAPR TC-12.
	The reason may be that, data in IAPR TC-12 is with much higher category number and feature dimensions, than that in MIRFlickr-25K and NUS-WIDE.
	\item As an online cross-modal hashing method, LCDH still can outperform some state-of-the-art offline cross-modal hashing methods in all cases.
	The main reason may be that, LCDH has powerful semantic preservation ability by fusing contrastive features extracted by CLIP, and the fused feature is further fed into the attention module to enhance the feature representation.
	\item LCDH has achieved obvious improvements on mAP results for both I$\rightarrow$T and T$\rightarrow$I tasks, especially when the length of hash codes is short.
	Notably, LCDH can deliver the remarkable and best results on mAP, of both I$\rightarrow$T and T$\rightarrow$I tasks for all three datasets.
\end{itemize}

\subsubsection{Results on Precision Recall Curve}
By measuring the relevance of the precision and retrieval results, precision recall curve can reveal the relationship between them.
Therefore, the higher the precision recall curve, the better the performance on retrieval.
From Figs. \ref{mirflickr_PR_TopN}, \ref{iaprtc_PR_TopN} and \ref{nuswide_PR_TopN}, we have the following observations and analyses of precision recall curve.
\begin{itemize}
	\item In all case, there are obvious gaps between the curves of the proposed LCDH and the second best baseline methods.
	This is consistent to the mAP results, indicating that LCDH can achieve competitive retrieval performance on precision in all levels of recall.
	\item All the methods including the proposed LCDH perform better on T$\rightarrow$I than I$\rightarrow$T tasks, in terms of the precision recall curve.
	The main reason may be that, the fusion of contrastive learning on cross-modal features can well extract the coexistent semantic relevance representation, which is further enhanced by the attention module for narrowing the heterogeneous gaps on semantic.
	\item In all levels of the curves, LCDH can deliver the highest precision consistently, for both I$\rightarrow$T and T$\rightarrow$I tasks on all three datasets.
	This indicates that LCDH is able to effectively capture the adequate coexistent semantic relevance, by the well designed feature extraction model and knowledge distillation framework in LCDH. 
\end{itemize}

\begin{table*}[!htbp]
	\centering
	\setlength{\tabcolsep}{1mm}
	\fontsize{9}{9}\selectfont
%	\fontsize{5}{4}\selectfont
	\begin{tabular}{ cccccccccccccccccc}
		\toprule[1.0pt]
		\multirow{2}{*}{Task}& \multirow{2}{*}{Variants} & \multicolumn{4}{c}{MIRFlickr-25K}   & \multicolumn{4}{c}{IAPR TC-12}   & \multicolumn{4}{c}{NUS-WIDE} 
		\\ 
		\cmidrule(r){3-6}  \cmidrule(r){7-10}  \cmidrule(r){11-14} 
		&                         & 16 bits          & 32 bits          & 64 bits          & 128 bits         & 16 bits          & 32 bits          & 64 bits          & 128 bits         & 16 bits          & 32 bits          & 64 bits          & 128 bits         \\ 
		\toprule[1.0pt]
		\multirow{8}{*}{\begin{tabular}[c]{@{}l@{}}   I$\rightarrow$T \\\\\\\\  \end{tabular}}  
		& $\textit{Log-Similarity --}$  & 0.7785 & 0.7813  & 0.7907  & 0.7981  & 0.5524  & 0.5838  & 0.6074  & 0.6253  & 0.7462  & 0.7697  & 0.7915  & 0.7991 \\
		& $\textit{Offline Hashing --}$  & 0.7609 & 0.7664  & 0.7699  & 0.7702  & 0.5389  & 0.5573  & 0.5897  & 0.6004  & 0.7277  & 0.7439  & 0.7741  & 0.7865 \\
		& $\textit{CLIP --}$  & 0.7671 & 0.7698  & 0.7710  & 0.7716  & 0.5427  & 0.5684  & 0.5913  & 0.6107  & 0.7324  & 0.7572  & 0.7890  & 0.7906  \\
		\cmidrule(r){3-6}  \cmidrule(r){7-10}  \cmidrule(r){11-14} 
		& \textbf{LCDH}      & \textbf{0.7898} & \textbf{0.7992} & \textbf{0.8034} & \textbf{0.8041}  &\textbf{0.5656} &\textbf{0.6093} &\textbf{0.6301} &\textbf{0.6499}  & \textbf{0.7564} & \textbf{0.7803}  & \textbf{0.8007} & \textbf{0.8015} \\ 
		\midrule[1.0pt]
		\multirow{9}{*}{\begin{tabular}[c]{@{}l@{}} T$\rightarrow$I \\\\\\\\\\ \end{tabular}}
		& $\textit{Log-Similarity --}$  & 0.8069 & 0.8257  & 0.8447  & 0.8532  & 0.5854  & 0.6043  & 0.6260  & 0.6598  & 0.7875  & 0.8011  & 0.8109  & 0.8212 \\
		& $\textit{Offline Hashing --}$  & 0.7943 & 0.8102  & 0.8228  & 0.8395  & 0.5753  & 0.5874  & 0.6183  & 0.6444  & 0.7790  & 0.7906  & 0.7975  & 0.8090 \\
		& $\textit{CLIP --}$  & 0.7998 & 0.8144  & 0.8316  & 0.8469  & 0.5791  & 0.5990  & 0.6247  & 0.6585  & 0.7816  & 0.7951  & 0.8002  & 0.8133  \\
		\cmidrule(r){3-6}  \cmidrule(r){7-10}  \cmidrule(r){11-14} 
		& \textbf{LCDH}    & \textbf{0.8223} & \textbf{0.8480} & \textbf{0.8556} & \textbf{0.8574}  &\textbf{0.5987} &\textbf{0.6229} &\textbf{0.6574} &\textbf{0.6810}  & \textbf{0.8018} & \textbf{0.8116}  & \textbf{0.8268} & \textbf{0.8311} \\ 
		\bottomrule[1.0pt]
	\end{tabular}
	\caption{The ablation results of LCDH on the three datasets with various code lengths.}
	\label{table_ablation_study}
\end{table*}

\subsubsection{Results on Top-N Precision Curve}
Normally, top-N precision curve can reveal the proportion of the top N results, which are with the highest probability in prediction results containing correct label.
Therefore, the higher the top-N precision curve, the better the performance on retrieval.
From Figs. \ref{mirflickr_PR_TopN}, \ref{iaprtc_PR_TopN} and \ref{nuswide_PR_TopN}, we can have the following observations and analyses of top-N precision curve.
\begin{itemize}
	\item There exists significant gaps between the top-N precision curves of the proposed LCDH and other baseline methods, especially on MIRFlickr-25K and IAPR TC-12.
	\item LCDH surpasses other baseline methods with a diverse numbers of retrieval instances, in terms of top-N precision on all three datasets.
	\item When the retrieval number of instance increasing largely, the proposed LCDH can still achieve the prominent and stable retrieval performance.
	This verifies that, LCDH can effectively and correctly transfer the coexistent semantic relevance information to student network by the similarity approximation and knowledge distillation framework proposed in LCDH.
\end{itemize}

\subsection{Ablation Analysis}
mAP results of the ablation analysis are represented in Table \ref*{table_ablation_study}, where '$\textit{--}$' means the corresponding module in the proposed LCDH is dropped and the optimization will be redid.

As shown in Table \ref*{table_ablation_study}, '$\textit{Offline Hashing}$' and '$\textit{CLIP}$' have considerable contributions to performances of LCDH, while '$\textit{Log-Similarity}$' have relatively weak contribution to the performance of LCDH.
The results of '$\textit{Offline Hashing}$' and '$\textit{CLIP}$' have verified the effectiveness of knowledge distillation framework and the CLIP-based feature extraction in the proposed LCDH.
That is, the deep semantic information is well extracted and preserved in teacher network, and then effectively transferred to the student network for guiding the online cross-modal hashing.
Moreover, the results also demonstrate that log-likelihood similarity preservation term is well designed to guide the training in student network.
Finally, the LCDH with full modules had achieved the best results in all cases, revealing that all the modules have indeed contributions to the performance of the proposed LCDH.

\subsection{Convergence Analysis}
\begin{figure}[!t]
	\centering
	\includegraphics[width=1.3in]{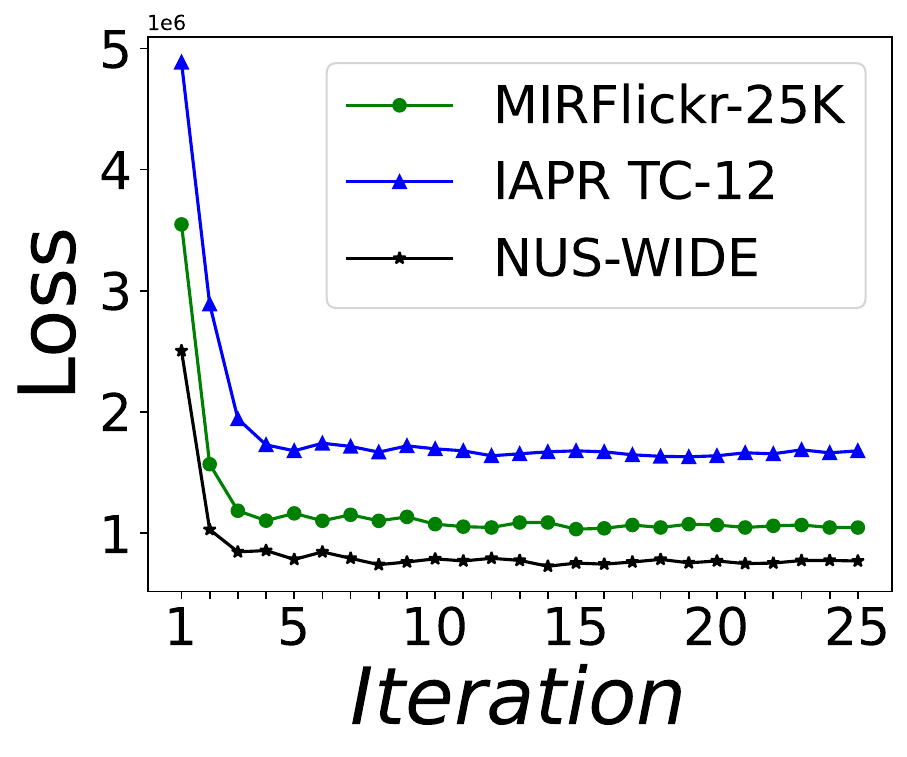}
	\centering
	\caption{The convergence analysis for LCDH on MIRFlickr-25K, IAPR TC-12 and NUS-WIDE @ 128 bits.}
	\label{ConvergenceAnalysis}
\end{figure}

By recording and illustrating the values of the objective function (i.e., Eq. (\ref*{objectiveFunction})), the convergence analysis for LCDH is provided as shown in Fig. \ref{ConvergenceAnalysis}.
From the figure, it can be observed easily that the losses (the values of the objective function) of LCDH are reducing quickly and becoming stable after around ten iterations for all datasets.
This verifies that LCDH can quickly become convergent and deliver a stable and consistent performance. 
That is, the quick convergence of losses can prove that LCDH effectively transfers the coexistent semantic relevance on modalities to guide the online hashing by the knowledge distillation framework.

\subsection{Parameter Sensitivity}
\begin{figure}[!t]
	\centering
	\subfigure[{\scriptsize Parameter sensitivity of $\lambda_{1}$ and $\lambda_{2}$ for I$\rightarrow$T tasks on MIRFlickr-25K}]{
		\includegraphics[width=1.4in]{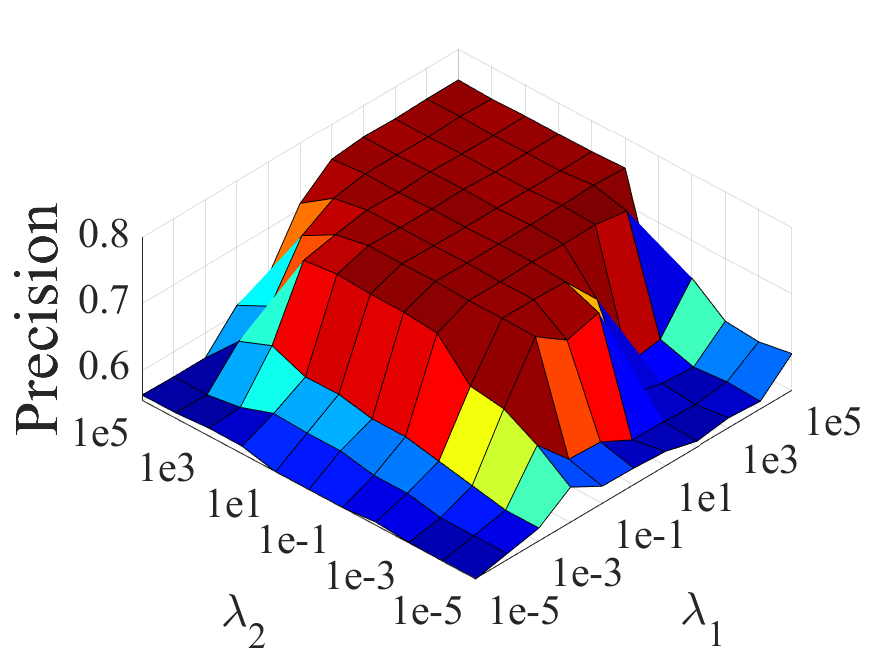}
		\label{parameterAnalysis_i2t_mir}
	}\hspace{.05in}
	\subfigure[{\scriptsize Parameter sensitivity of $\lambda_{1}$ and $\lambda_{2}$ for T$\rightarrow$I tasks on MIRFlickr-25K}]{
		\includegraphics[width=1.4in]{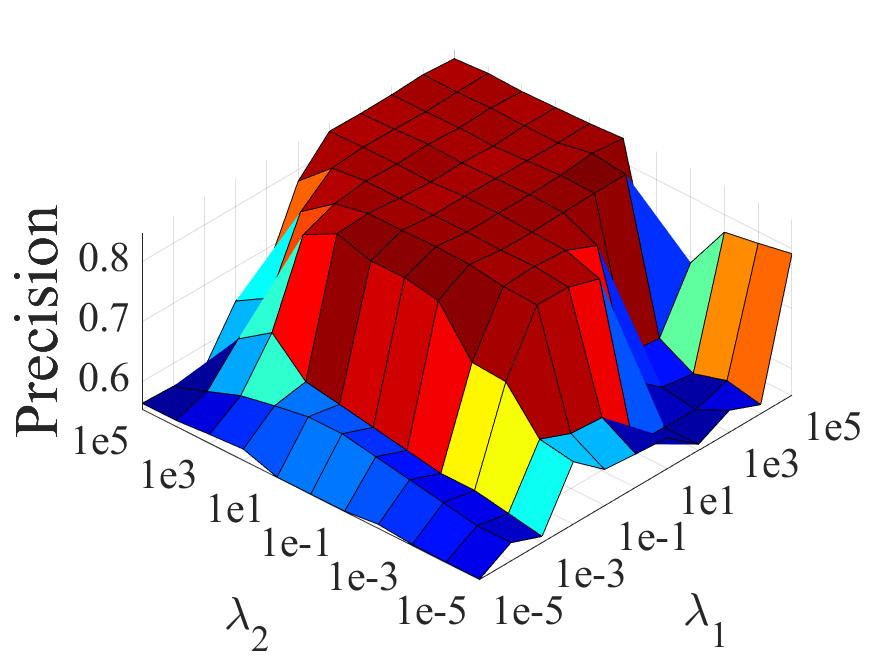}
		\label{parameterAnalysis_t2i_mir}
	}
	\centering
	\caption{The parameter analysis for LCDH of $\lambda_{1}$ and $\lambda_{2}$, for I$\rightarrow$T and T$\rightarrow$I tasks on MIRFlickr-25K @ 128 bits.}
	\label{ParameterAnalysis}
\end{figure}

In the objective function of LCDH, parameter $\lambda_1$ controls the impact of the similarity preservation in teacher network, while parameter $\lambda_2$ controls the impact of similarity approximation for offline and online similarity in knowledge distillation.
The ranges of $\lambda_{1}$ and $\lambda_{2}$ are set to \{$1e^{-5}, 1e^{-4}, 1e^{-3}, 1e^{-2}, 1e^{-1}, 1e^{0}, 1e^{1}, 1e^{2}, 1e^{3}, 1e^{4}, 1e^{5}$\}.
As shown in Fig. \ref{ParameterAnalysis}(a) and Fig. \ref{ParameterAnalysis}(b), LCDH achieved the best performance on MIRFlickr-25K for both I$\rightarrow$T and T$\rightarrow$I retrieval tasks when $\lambda_{1}=1e^{4}$, $\lambda_{2}=1e^{0}$.
From the figures, we also can find that both $\lambda_{1}$ and $\lambda_{2}$ have make contributions to the performances of LCDH to great extents.
It indicates that the deep feature extraction, representation enhancement, similarity approximation and the knowledge distillation framework are effectively designed in LCDH.

%\columnwidth

\section{Conclusion}
This paper proposes a lightweight contrastive distilled hashing (LCDH) method for online cross-modal retrieval. 
LCDH fuses CLIP-extracted cross-modal features and enhances representations using an attention module. 
Binary codes with label information are generated via a fully connected layer to align offline and online similarity matrices. 
In the student network, image and text features from VGG16 and BoW are similarly transformed into binary codes. 
By approximating similarities and leveraging semantic relevance from the teacher network, LCDH effectively bridges offline and online hashing, achieving competitive retrieval performance with state-of-the-art offline methods, while maintaining a lightweight structure.

\section{Acknowledgments}
This work was supported in part by the National Natural Science Foundation of China under Grant 62302112, Grant 62006048, and Grant 62176065; in part by the Guangdong Pearl River Talent Program under Grant  2023QN10X503; in part by the Guangzhou Basic and Applied Basic Research Foundation under Grant 2025A04J3378; in part by the Guangdong
Provincial National Science Foundation under Grant 2021A1515012017; and in part by the Guangzhou University Research Project under Grant RQ2021013.

\bibliography{aaai25}

\end{document}